\documentclass{article}
\usepackage{mlspconf, graphicx}
\usepackage{amsmath,amssymb,amsfonts}
\usepackage{booktabs}
\usepackage{cite}
\usepackage{comment}
\usepackage{balance}
\usepackage{url}

\usepackage{arydshln}

\copyrightnotice{979-8-3503-2411-2/23/\$31.00 {\copyright}2023 IEEE}

\toappear{2023 IEEE International Workshop on Machine Learning for Signal Processing, Sept.\ 17--20, 2023, Rome, Italy}

\title{PHYDI: Initializing Parameterized Hypercomplex Neural Networks as Identity Functions}

\name{Matteo Mancanelli, Eleonora Grassucci, Aurelio Uncini, and Danilo Comminiello \thanks{Corresponding author's email: eleonora.grassucci@uniroma1.it. We acknowledge financial support from PNRR MUR project PE0000013-FAIR.}}
\address{Dept. of Information Engineering, Electronics, and Telecomm., Sapienza University of Rome, Italy}

\begin{document}

\maketitle

\begin{abstract}
Neural models based on hypercomplex algebra systems are growing and prolificating for a plethora of applications, ranging from computer vision to natural language processing. Hand in hand with their adoption, parameterized hypercomplex neural networks (PHNNs) are growing in size and no techniques have been adopted so far to control their convergence at a large scale. In this paper, we study PHNNs convergence and propose parameterized hypercomplex identity initialization (PHYDI), a method to improve their convergence at different scales, leading to more robust performance when the number of layers scales up, while also reaching the same performance with fewer iterations. We show the effectiveness of this approach in different benchmarks and with common PHNNs with ResNets- and Transformer-based architecture. The code is available at \url{https://github.com/ispamm/PHYDI}.
\end{abstract}


\section{Introduction}
\label{sec:intro}

\begin{figure}
    \centering
    \includegraphics[width=\linewidth]{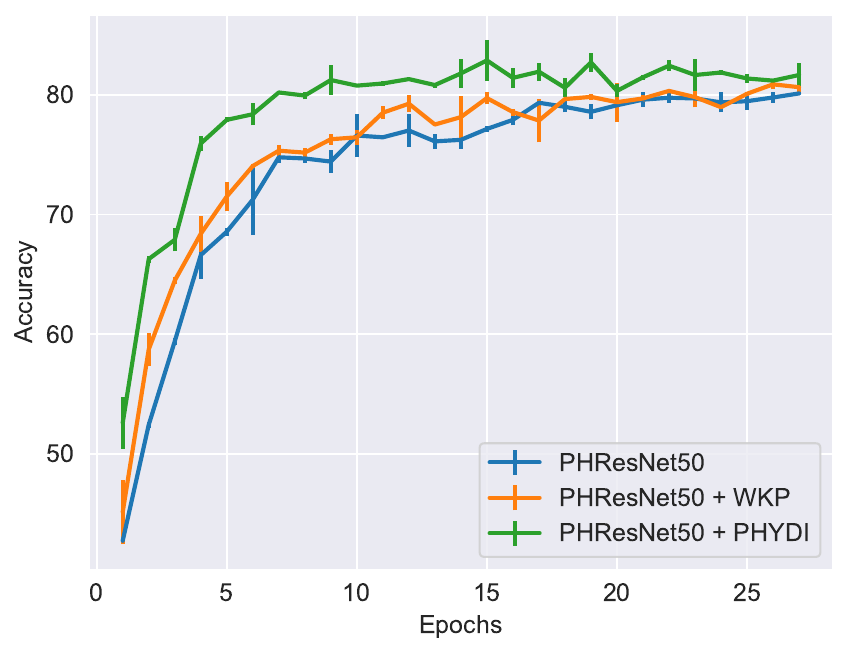}
    \caption{The proposed PHYDI initialization speeds up the convergence of parameterized hypercomplex neural networks in which it is employed.}
    \label{fig:res_convergence}
\end{figure}

Although the largest part of deep learning models is defined following the rules of real-valued numbers and algebra $\mathbb{R}$, such a choice is not always the best one for multidimensional data. Therefore, an increasing number of works are exploring the possibility of defining models with different underlying algebras that better fit the problem of study. Among these, Clifford, Cayley-Dickson, and hypercomplex algebras have been widely adopted. More recently, Parameterized hypercomplex neural networks (PHNNs) have transformed the wide research field of neural models defined over such hypercomplex algebras. This happened thanks to their flexibility and malleability that allow users to make use of hypercomplex-based networks without developing ad-hoc architectures or seeking the proper algebra domain that best fits the specific task. Indeed, PHNNs grasp algebra rules directly from data and therefore they can be employed for any $n$-dimensional data without modifications to architectural layers. For that reason, the already ample field of hypercomplex models based on complex \cite{Trabelsi2017DeepCN}, quaternion \cite{Jia2022TIP}, dual quaternion \cite{Poppelbaum2022Access, grassucci23DualQ}, and octonion \cite{Trabelsi2017DeepCN} numbers has been permeated by PHNNs. These networks have been defined with different known backbones such as ResNets \cite{grassucci2021phnns, lopez2022hypercomplex}, GANs \cite{Sfikas2022Springer, Grassucci2022IJCNN}, graph neural networks \cite{Le2021ParameterizedHG}, and Transformers \cite{zhang2021phm}, among others \cite{Basso2022TowardsEE, Panagos2022Compress}.
So as with any neural model, their expressiveness increases with deeper representations, also due to the fact that for high values of the hyperparameter $n$, which also affects the number of parameters reducing them to $1/n$, PHNNs may be defined with very few parameters even if the number of layers is large.

However, any convergence study or regularization and normalization strategies have been proposed for improving PHNNs training stability and convergence when the number of layers increases. Indeed, PHNNs behavior with a large number of layers is still unknown, as well as it is not clear how the parameters reduction driven by the hyperparameter $n$ affects the learning of very deep networks and whether intra-layer parameters and overall parameters are balanced during training.

In this paper, therefore, we first conduct a study on PHNNs convergence in large-scale training, discovering that very deep architectures have convergence issues, and founding that the hyperparameter $n$ is related to the convergence.
In order to address these issues, we propose parameterized hypercomplex identity initialization (PHYDI), a method to help PHNNs converge fast and improve large-scale model learning motivated by the dynamical isometry that has been proved a clear indicator of trainability \cite{Pennington2017ResurrectingTS}. We propose to initialize each parameterized hypercomplex multiplication (PHM) or convolution (PHC) layer, that represent the core of PHNNs, as an identity function. The proposed initialization is carried out by adding a residual connection and a trainable zero-initialized parameter that multiplies the actual layer parameters, as introduced for real-valued models \cite{ReZero2021PMLR}.

We prove that PHYDI improves very deep ResNets- and Transformer-based PHNNs convergence in different benchmarks even when standard PHNNs diverge, therefore improving the learning ability of large architectures. Furthermore, the proposed method leads to faster convergence of each PHNN we test, both in image and language datasets. 

In summary, our contributions are: i) We conduct, to the best of our knowledge, the first study on the convergence of PHNNs in large-scale training, showing that this is also related to the key hyperparameter $n$. ii) We propose PHYDI, a method to avoid divergence of very deep PHNNs, which also fastens the convergence of convolutional- and attention-based PHNNs in multiple benchmarks, allowing the learning of large-scale networks with fewer iterations.

The rest of the paper is organized as follows. The background on PHNNs is developed in Section \ref{sec:back}, the proposed method is presented in Section \ref{sec:method}, while the experimental evaluation is performed in Section \ref{sec:exp}. Finally, conclusions are drawn in Section \ref{sec:conc}.

\section{Parameterized Hypercomplex Neural Networks}
\label{sec:back}


Although the largest part of neural models is defined over the set of real numbers $\mathbb{R}$, this configuration does not always perfectly fit every task. In the case of multidimensional data, such as objects in the 3D space, multichannel audio signals, or color images, real-valued models break the multidimensional nature of the input and their learning may fail. For this reason, recently, several works are considering other numerical domains with stronger algebraic and geometry properties that better model the multidimensional structure of the data. Among these domains, the complex (one imaginary unit) and the quaternion (three imaginary units) ones have been the most cited due to their 2D and 4D nature and their algebraic properties that can be leveraged during the learning phase. Indeed, the so-called Hamilton products of quaternions allow a parameter reduction up to the $75\%$ and, most importantly, the parameters sharing within the layer. This feature enables the model equipped with such algebra to preserve local relations and correlations among the input dimensions, building a more accurate view of the multidimensional data. However, such models are limited to domain dimensionality and therefore arduous to apply when data does not fit the domain-specific dimensions, such as 2 for the complex domain or 4 for the quaternion one. For this reason, parameterized hypercomplex neural networks (PHNNs) have been introduced in the literature \cite{grassucci2021phnns}. The family of PHNNs comprises neural models defined over various domains by means of hypercomplex layers parameterized by the user. The core operation of these networks is the parameterized hypercomplex (PH) layer that builds the weight matrix as a sum of Kronecker products as follows:

\begin{equation}
    \mathbf{H} = \sum_{i=1}^n \mathbf{A}_i \otimes \mathbf{F}_i,
\end{equation}

\noindent where the hyperparameter $n$ defines the domain in which the model operates (e.g., $n=2$ complex domain, $n=4$ quaternion domain, and so on), and, most importantly, it can be set to any integer value even though the algebra regulations are not known. That is possible thanks to the matrices $\mathbf{A}_i$ that learn the algebra rules directly from data, and the matrices $\mathbf{F}_i$ representing the parameters (or filters for convolutional layers).
Thanks to their flexibility, PH layers can then be involved in several common neural architectures such as ResNets \cite{grassucci2021phnns}, transformers \cite{zhang2021phm}, and graph neural networks \cite{Le2021ParameterizedHG}, just as a replacement for standard real-valued layers. Although PHNNs show outstanding results in several tasks, such models are tremendously new in the literature and their convergence behavior has not been studied yet. 

\section{PHYDI: Parameterized hypercomplex identity initialization}
\label{sec:method}

\begin{figure}
    \centering
    \includegraphics[width=\linewidth]{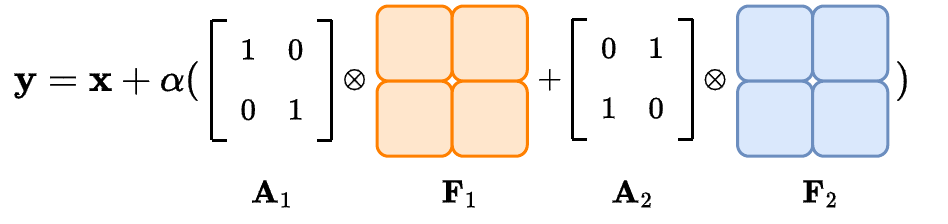}
    \caption{A simple PHM formulation for the proposed method with $n=2$. The PHM layer builds the weights matrix $\mathbf{H}$ as a sum of $2$ Kronecker products that are multiplied by the PHYDI learnable parameter $\alpha$, initialized to $0$ so as to ensure the identity function brought by the inserted residual connection. Note that the matrices $\mathbf{A}_1$ and $\mathbf{A}_2$ are learnable during training, exactly as the classical weights matrix $\mathbf{F}_1$ and $\mathbf{F}_2$.}
    \label{fig:phydi}
\end{figure}

\begin{figure}
    \centering
    \includegraphics[width=0.8\linewidth]{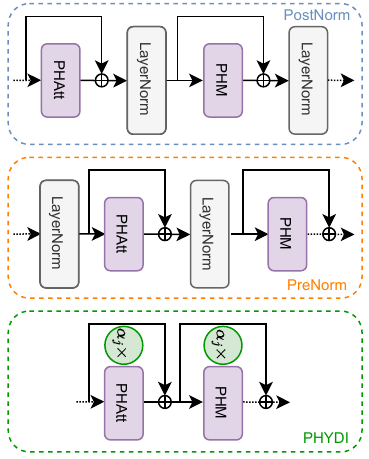}
    \caption{The proposed PHYDI formulation for PHTransformer layers compared with common PostNorm \cite{vaswani2017attention} and PreNorm \cite{Xiong2020OnLN} architectures.}
    \label{fig:phtransf}
\end{figure}


The proposed parameterized hypercomplex identity initialization (PHYDI) for PHNNs is easy to be involved in any pre-existing PHNN as it consists of a slight modification to the hypercomplex architecture to perform the identity operation. Through identity, we can ensure initial dynamical isometry that has been proven to be a clear aspect of well-trainable networks \cite{Pennington2017ResurrectingTS}.
In practice, to simplify the gradient propagation at the initialization, the signal should not propagate on the PH layer $\mathcal{F}\{\mathbf{H}\}$, but rather on its residual connection $\mathbf{x}$. To do that, a parameter $\alpha$ is set to multiply the PH layer and initialized to $0$, so that only the residual connection remains active during the first iteration. More formally, the $j+1\text{-th}$ PH layer with PHYDI formulation becomes:

\begin{equation}
\label{eq:rezero}
    \mathbf{x}_{j+1} = \mathbf{x} + \alpha_j \mathcal{F}\{\mathbf{H}\}(\mathbf{x}),
\end{equation}

\noindent whereby $\alpha$ is a learnable parameter initialized to $0$ at the beginning of the training. A visual representation is shown in Fig. \ref{fig:phydi}, where the Kronecker product blocks that build the PH layer are then summed and multiplied by the PHYDI learnable parameter $\alpha$.
In the next subsections, we formally define two of the most common vision and language PH architectures with the proposed PHYDI method, which are PHResNets and PHTransformers.

\subsection{Initialization of Hypercomplex ResNets}

PHResNets have already been defined with residual connections, so no architectural changes are needed except for the insertion of the PHYDI parameter $\alpha$ that initializes the network to perform the identity operation. Therefore, the PHResNet layer will be defined exactly as \eqref{eq:rezero}, where instead of generic layers, the function $\mathcal{F}(\mathbf{x}, \{ \mathbf{H}_j \})$ comprises parameterized hypercomplex convolutional (PHC) layers:

\begin{equation}
    \mathcal{F}(\mathbf{x}, \{ \mathbf{H}_j \}) = \text{PHC} \left( \text{ReLU} \left( \text{PHC}(\mathbf{x} ) \right) \right).
\end{equation}

Although this formulation is already based on residual connections, we will demonstrate that the identity initialization due to the $\alpha$ parameter gives PHResNets a faster convergence, especially in very deep networks.

\subsection{Initialization of Hypercomplex Transformers}

PHTransformers also have a residual connection structure inside their layers. However, its composition is more complex than PHResNets one and a more detailed study has to be performed for the identity initialization. Indeed, a standard PHTransformer layer can be described by

\begin{equation}
    \begin{split}
    \mathbf{x}_{j+1} = &\text{ LayerNorm} \{ \mathbf{x}_j + \text{PHM}( \\
    &\text{ LayerNorm}(\mathbf{x}_j + \text{PHAtt}(\mathbf{x}_j)) \},
    \end{split}
\end{equation}

\noindent that is, a post-normalization of the sub-layer modules (Parameterized hypercomplex multi-head attention (PHAtt) and Parameterized hypercomplex multiplication (PHM)). Following the literature suggestion \cite{ReZero2021PMLR}, in order to initialize the layer as the identity function, we can remove the layer normalization and insert the PHYDI parameters as multipliers for the sub-layers as follows:

\begin{equation}
    \mathbf{x}_{j+1} = \mathbf{x}_j + \alpha_j \text{PHM}(
    \mathbf{x}_j + \alpha_j \text{Att}(\mathbf{x}_j)).
\end{equation}

Note that the learnable parameter $\alpha_j$ is the same within the PHTransformer layer and it is initialized to $\alpha_j=0$ to ensure the identity operation. Figure \ref{fig:phtransf} shows the three different PHTransformer configurations we test in our experiments, namely PostNorm \cite{vaswani2017attention}, PreNorm \cite{Xiong2020OnLN}, and PHYDI (proposed).

\section{Experiments}
\label{sec:exp}

\begin{table}[t]
\caption{PHResNets with standard, WKP, and PHYIDI initialization for different values of the hyperparameter $n$ in the CIFAR10 dataset. Metrics M1: Epochs to $80\%$ Acc, M2: \# Epochs to beat one w/ PHYDI. The uncertainties correspond to standard error.}
\vspace{0.2cm}
\label{tab:res}
\centering
\begin{tabular}{@{}lccc@{}}
\toprule
\multicolumn{1}{c}{Model} & $n$ & M1$\downarrow$ & M2 \\ \midrule
PHResNet18      & 2     & 6.00 $\pm$ 0.58  & 2 \\
\quad + WKP    & 2     & \textbf{5.75} $\pm$ 0.25  & 2 \\
\quad + PHYDI   & 2     & 6.00 $\pm$ 0.00  & - \\
PHResNet50      & 2     & 10.67 $\pm$ 1.20          & 3 \\
\quad + WKP    & 2     & 10.67 $\pm$ 0.67          & 2 \\
\quad + PHYDI   & 2     & \textbf{7.00} $\pm$ 0.58  & - \\
PHResNet152     & 2     & 32.67 $\pm$ 2.03          & 4 \\
\quad + WKP    & 2     & 29.80 $\pm$ 3.68          & 4 \\
\quad + PHYDI   & 2     & \textbf{6.33} $\pm$ 1.33  & - \\ \midrule
PHResNet18      & 3     & 6.33 $\pm$ 0.33           & 2 \\
\quad + WKP    & 3     & 6.00 $\pm$ 0.00           & 1 \\
\quad + PHYDI   & 3     & \textbf{5.00} $\pm$ 0.58  & - \\
PHResNet50      & 3     & 8.67 $\pm$ 1.20           & 2 \\
\quad + WKP    & 3     & 9.0 $\pm$ 0.58            & 2 \\
\quad + PHYDI   & 3     & \textbf{6.33} $\pm$ 0.67  & - \\
PHResNet152     & 3     & 26.67 $\pm$ 1.76          & 4 \\
\quad + WKP    & 3     & 20.00 $\pm$ 2.12          & 4 \\
\quad + PHYDI   & 3     & \textbf{4.67} $\pm$ 0.33  & - \\ \midrule
PHResNet18      & 4     & \textbf{3.75} $\pm$ 0.48  & 1 \\
\quad + WKP    & 4     & 5.67 $\pm$ 0.67           & 2 \\
\quad + PHYDI   & 4     & 4.50 $\pm$ 0.50           & - \\
PHResNet50      & 4     & 8.33 $\pm$ 0.88           & 2 \\
\quad + WKP    & 4     & 9.33 $\pm$ 0.88           & 2 \\
\quad + PHYDI   & 4     & \textbf{7.00} $\pm$ 1.15  & - \\
PHResNet152     & 4     & 22.67 $\pm$ 3.71          & 3 \\
\quad + WKP    & 4     & 20.60 $\pm$ 1.60          & 3 \\
\quad + PHYDI   & 4     & \textbf{5.33} $\pm$ 0.33  & - \\
 \bottomrule
\end{tabular}
\end{table}

In this Section, we present the experimental validation of our theoretical claims. We conduct experiments on two common tasks such as image classification and sequence-to-sequence prediction. Therefore, we involve different ResNets backbones (ResNet18, ResNet50, and ResNet152) for the first task, and a Transformer-based model varying the number of layers for the second task.

\subsection{Comparison methods}

For convolutional networks, no previous initialization or fast-convergence method have been proposed for PHHNs, so we compare our proposal with standard PHResNets with classical initialization \cite{grassucci2021phnns}. Additionally, we experiment with Fixup initialization \cite{Zhang2019Fixup}, but we note that it required a senseless number of epochs for reaching the $80\%$ of accuracy with PHResNets18 and almost damaged the final accuracy performances for deeper PHResNets. Moreover, identity initialization methods have been already proven to be better than Fixup for real-valued models \cite{ReZero2021PMLR}. For these reasons, we do not report Fixup scores in Table \ref{tab:res}. We further validate PHYDI against a weighted non-identity initialization of the Kronecker products. We insert a learnable weight $\alpha$ initialized to $1$ in each PH layer product in order to let the network learn the proper weighting mechanism for each Kronecker multiplication, and therefore, for each input dimension. However, the weighted Kronecker product (WKP) does not ensure identity initialization.

Regarding transformers, we compare PHTransformers with different layer normalization positions, such as the original real-valued Transformer PostNorm \cite{vaswani2017attention}, the newer PreNorm \cite{Xiong2020OnLN}, and the proposed PHYDI initialization without normalizations.

\subsection{Results on faster convergence}

Table \ref{tab:res} shows the image classification results for different PHResNets with and without the proposed PHYDI method. We conduct experiments with three different ResNets backbones with an increasing number of layers (18, 50, 152) and for different configurations of the hyperparameter $n$. We train all the models following the recipe of the original real-valued ResNet paper \cite{he2016residual} that the PHResNet paper has proved to work well also in hypercomplex domains \cite{grassucci2021phnns}. We just edit the batch size, using a value of $64$. We experiment on the image classification task with the CIFAR10 dataset in order to test the method on a standard benchmark. 
We evaluate the performances of our initialization method according to two metrics, namely the number of epochs the models require to reach the $80\%$ of accuracy (M1), and the number of epochs to beat a model with the proposed approach (M2).
PHYDI ensures faster convergence in every test we conduct, lowering the number of epochs to obtain the $80\%$ of accuracy. Furthermore, the advantages of our approach are shared among the most common choices of the hyperparameters $n$, proving that PHYDI PHResNets are robust across architectural changes.
Moreover, the effect of the proposed method is more evident with the increase of model layers. Indeed, while PHResNets152 suffers from slow convergence, endowing such networks with PHYDI initialization drastically fastens the convergence, lowering the number of epochs by more than $20$ in some cases. This is clear evidence of the incoming results regarding PHYDI ability in allowing improved learning of large-scale models.

\begin{figure*}
    \centering
    \includegraphics[width=\textwidth]{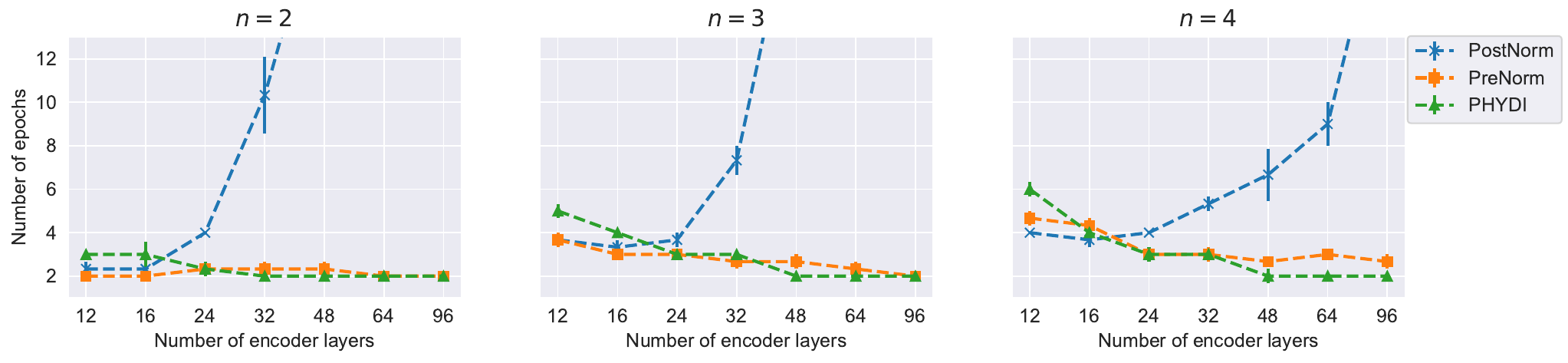}
    \caption{Number of epochs to reach a perplexity value $\ge 200$ in the WikiText2 dataset for PH Transformers with increasing depth of the encoder model from $12$ to $96$. The three plots refer to different values of the hyperparameter $n=2,3,4$. Vertical standard error bars are shown too, computed over multiple runs. The PH Transformer equipped with PHYDI preserves the performance even in large-scale models while standard and PostNorm models diverge.}
    \label{fig:perplexity}
\end{figure*}

\subsection{Results on large-scale models}

The signal and gradient propagation in real- as well as in hypercomplex-valued neural networks becomes much more difficult as the model depth increases. Therefore, proper forward and backward flows are crucial for the effective learning of a neural model. In this subsection, we focus our experiment on the sequence-to-sequence PHTransformers \cite{zhang2021phm} on the WikiText2 dataset. As in the previous tests, we experiment PHNNs with different values of the hyperparameter $n=2,3,4$.
We train all models for $50$ epochs with a batch size of $64$ and optimize them with Adagrad and a step learning rate scheduler that progressively reduces it from its original value of $0.01$. We experiment with different encoder depths and a number of layers in the set $[12, 16, 24, 32, 48, 64, 96]$. Similar to previous experiments, we set up a metric to prove the fast convergence: we consider the number of epochs to reach a value of perplexity equal to $200$.

Figure \ref{fig:perplexity} shows the results of our experiments for different $n$ settings. The first consideration we can draw is that the original Transformer architecture defined in hypercomplex domains with PostNorm diverges as the number of encoder layers increases. This result extends real-valued results already proven in \cite{ReZero2021PMLR}. However, this configuration seems linked to the number of parameters of the model and, consequently, to the choice of the hyperparameter $n$ in PHNNs. Indeed, it is interesting to note that while for $n=2$ the PostNorm configuration diverges with just $48$ layers, acceptable results are instead obtained up to $64$ layers when setting $n=4$. Therefore, the gradient is better propagated with fewer parameters and model convergence does not depend only on the depth of the model itself. 

The PreNorm method performs better than the original PostNorm, providing stabler performances even when the depth increases. Additionally, this method performs best for small encoders with $12$ or $16$ layers, exceeding also the proposed approach. However, PHYDI clearly improves PHTransformer convergence and robustness when the architecture becomes very deep, especially from $32$ layers up to the maximum of our experiments. The PHTransformer equipped with PHYDI initialization still requires just $2$ epochs to reach a perplexity value of $200$ even with $96$ encoder layers, proving that the proposed approach improves the learning of large-scale PHNNs. Overall, it is important to note that PHYDI improves large models stability and it also provides robust results also for unfavorable configurations, e.g. small networks.

\subsection{Robustness to hyperparameters settings}

We perform additional experiments with different hyperparameters settings further to validate PHYDI and its robustness to various configurations. In detail, we conduct tests varying the learning rate, considering both $0.01$ and $0.1$, and changing the batch size (from $64$ to $128$), since this usually affects convergence speed. Table \ref{tab:hyper} shows the results of the experiments with PHResNets18 and PHResNets50 with the same evaluation metrics as Table \ref{tab:res}. It is evident that PHYDI improves results under different settings too, being robust to various hyperparameters configurations.

\begin{table}[t]
\caption{PHResNets results with different initialization methods and learning rate configurations. The batch size is different from previous experiments. Results show that PHYIDI is robust across various settings, improving convergence speed.}
\label{tab:hyper}
\vspace{0.2cm}
\centering
\begin{tabular}{@{}lccccc@{}}
\toprule
\multicolumn{1}{c}{Model} & $n$ & btc-sz & lr & M1$\downarrow$ & M2 \\ \midrule
PHResNet18      & 2     & 128   & 0.1   & 21.86 $\pm$ 3.14  & 3 \\
\quad + WKP     & 2     & 128   & 0.1   & 17.17 $\pm$ 3.35  & 3 \\
\quad + PHYDI  & 2     & 128   & 0.1   & \textbf{10.00} $\pm$ 1.22  & - \\ \hdashline
PHResNet50      & 2     & 128   & 0.01  & 24.33 $\pm$ 2.19  & 3 \\
\quad + WKP     & 2     & 128   & 0.01  & 12.67 $\pm$ 1.67  & 2 \\
\quad + PHYDI  & 2     & 128   & 0.01  & \textbf{8.67} $\pm$ 0.88   & - \\ \midrule
PHResNet18      & 3     & 128   & 0.1   & 20.50 $\pm$ 5.08  & 3 \\
\quad + WKP     & 3     & 128   & 0.1   & 12.83 $\pm$ 1.68  & 3 \\
\quad + PHYDI  & 3     & 128   & 0.1   & \textbf{8.00} $\pm$ 0.58   & - \\ \hdashline
PHResNet50      & 3     & 128   & 0.01  & 11.25 $\pm$ 2.56  & 2 \\
\quad + WKP     & 3     & 128   & 0.01  & 12.80 $\pm$ 3.89  & 2 \\
\quad + PHYDI  & 3     & 128   & 0.01  & \textbf{10.75} $\pm$ 2.25  & - \\ \midrule
PHResNet18      & 4     & 128   & 0.1   & 21.40 $\pm$ 3.22  & 3 \\
\quad + WKP     & 4     & 128   & 0.1   & 20.67 $\pm$ 2.78  & 2 \\
\quad + PHYDI  & 4     & 128   & 0.1   & \textbf{8.80} $\pm$ 1.24   & - \\ \hdashline
PHResNet50      & 4     & 128   & 0.01  & 22.33 $\pm$ 8.67  & 2 \\
\quad + WKP     & 4     & 128   & 0.01  & 11.00 $\pm$ 0.58  & 2 \\
\quad + PHYDI  & 4     & 128   & 0.01  & \textbf{10.00} $\pm$ 1.00  & - \\ \bottomrule
\end{tabular}
\end{table}


\section{Conclusion}
\label{sec:conc}

In this paper, we study the convergence of common PHNNs, proposing parameterized hypercomplex identity initialization (PHYDI), a method that, with very few architectural changes, improves convergence in speed and learning depth. We experiment with PH convolutional and transformer models on known benchmarks and we prove that PHNNs endowed with PHYDI gain convergence speed and a better gradient propagation when the number of layers increases.

\balance
\bibliographystyle{IEEEtran}
\bibliography{PHZero.bib}

\begin{thebibliography}{10}
\providecommand{\url}[1]{#1}
\csname url@samestyle\endcsname
\providecommand{\newblock}{\relax}
\providecommand{\bibinfo}[2]{#2}
\providecommand{\BIBentrySTDinterwordspacing}{\spaceskip=0pt\relax}
\providecommand{\BIBentryALTinterwordstretchfactor}{4}
\providecommand{\BIBentryALTinterwordspacing}{\spaceskip=\fontdimen2\font plus
\BIBentryALTinterwordstretchfactor\fontdimen3\font minus
  \fontdimen4\font\relax}
\providecommand{\BIBforeignlanguage}[2]{{%
\expandafter\ifx\csname l@#1\endcsname\relax
\typeout{** WARNING: IEEEtran.bst: No hyphenation pattern has been}%
\typeout{** loaded for the language `#1'. Using the pattern for}%
\typeout{** the default language instead.}%
\else
\language=\csname l@#1\endcsname
\fi
#2}}
\providecommand{\BIBdecl}{\relax}
\BIBdecl

\bibitem{Trabelsi2017DeepCN}
C.~Trabelsi, O.~Bilaniuk, D.~Serdyuk, S.~Subramanian, J.~F. Santos, S.~Mehri,
  N.~Rostamzadeh, Y.~Bengio, and C.~Pal, ``Deep complex networks,'' \emph{ArXiv
  preprint arXiv:1705.09792}, 2017.

\bibitem{Jia2022TIP}
Z.~Jia, Q.~Jin, M.~K. Ng, and X.-L. Zhao, ``Non-local robust quaternion matrix
  completion for large-scale color image and video inpainting,'' \emph{IEEE
  Trans. on Image Process.}, vol.~31, pp. 3868--3883, 2022.

\bibitem{Poppelbaum2022Access}
J.~Pöppelbaum and A.~Schwung, ``Predicting rigid body dynamics using dual
  quaternion recurrent neural networks with quaternion attention,'' \emph{IEEE
  Access}, vol.~10, pp. 82\,923--82\,943, 2022.

\bibitem{grassucci23DualQ}
E.~Grassucci, G.~Mancini, C.~Brignone, A.~Uncini, and D.~Comminiello, ``Dual
  quaternion ambisonics array for six-degree-of-freedom acoustic
  representation,'' \emph{Pattern Recognition Letters}, vol. 166, pp. 24--30,
  Feb 2023.

\bibitem{grassucci2021phnns}
E.~Grassucci, A.~Zhang, and D.~Comminiello, ``{PHNN}s: Lightweight neural
  networks via parameterized hypercomplex convolutions,'' \emph{IEEE Trans. on
  Neural Netw. and Learning Systems}, pp. 1--13, dec 2022.

\bibitem{lopez2022hypercomplex}
E.~Lopez, E.~Grassucci, M.~Valleriani, and D.~Comminiello, ``Hypercomplex
  neural architectures for multi-view breast cancer classification,''
  \emph{arXiv preprint arXiv:2204.05798}, 2022.

\bibitem{Sfikas2022Springer}
G.~Sfikas, G.~Retsinas, B.~Gatos, and C.~Nikou, ``Hypercomplex generative
  adversarial networks for lightweight semantic labeling,'' in \emph{Pattern
  Recognition and Artificial Intelligence}.\hskip 1em plus 0.5em minus
  0.4em\relax Springer International Publishing, 2022, pp. 251--262.

\bibitem{Grassucci2022IJCNN}
E.~Grassucci, L.~Sigillo, A.~Uncini, and D.~Comminiello, ``Hypercomplex
  image-to-image translation,'' in \emph{Int. Joint Conf. on Neural Netw.
  (IJCNN)}, 2022.

\bibitem{Le2021ParameterizedHG}
T.~Le, M.~Bertolini, F.~No'e, and D.~A. Clevert, ``Parameterized hypercomplex
  graph neural networks for graph classification,'' in \emph{Int. Conf. on
  Artificial Neural Netw. (ICANN)}, 2021.

\bibitem{zhang2021phm}
A.~Zhang, Y.~Tay, S.~Zhang, A.~Chan, A.~T. Luu, S.~C. Hui, and J.~Fu, ``Beyond
  fully-connected layers with quaternions: Parameterization of hypercomplex
  multiplications with $1/n$ parameters,'' in \emph{Int. Conf. on Machine
  Learning ({ICML})}, 2021.

\bibitem{Basso2022TowardsEE}
L.~Basso, Z.~Ren, and W.~Nejdl, ``Towards efficient ecg-based atrial
  fibrillation detection via parameterised hypercomplex neural networks,''
  \emph{ArXiv preprint arXiv:2211.02678}, 2022.

\bibitem{Panagos2022Compress}
I.~I. Panagos, G.~Sfikas, and C.~Nikou, ``Compressing audio visual speech
  recognition models with parameterized hypercomplex layers,'' in
  \emph{Hellenic Conference on Artificial Intelligence}, ser. SETN '22, 2022.

\bibitem{Pennington2017ResurrectingTS}
J.~Pennington, S.~Schoenholz, and S.~Ganguli, ``Resurrecting the sigmoid in
  deep learning through dynamical isometry: theory and practice,'' in
  \emph{Advances in Neural Information Processing (NIPS)}, 2017.

\bibitem{ReZero2021PMLR}
T.~Bachlechner, B.~P. Majumder, H.~Mao, G.~Cottrell, and J.~McAuley, ``Rezero
  is all you need: fast convergence at large depth,'' in \emph{Conf. on
  Uncertainty in Artificial Intelligence}, ser. Proceedings of Machine Learning
  Research, vol. 161, 2021, pp. 1352--1361.

\bibitem{vaswani2017attention}
A.~Vaswani, N.~Shazeer, N.~Parmar, J.~Uszkoreit, L.~Jones, A.~N. Gomez,
  {\L}.~Kaiser, and I.~Polosukhin, ``Attention is all you need,'' in
  \emph{Advances in Neural Information Processing Systems (NeurIPS)}, 2017, pp.
  5998--6008.

\bibitem{Xiong2020OnLN}
R.~Xiong, Y.~Yang, D.~He, K.~Zheng, S.~Zheng, C.~Xing, H.~Zhang, Y.~Lan,
  L.~Wang, and T.-Y. Liu, ``On layer normalization in the transformer
  architecture,'' in \emph{Int. Conf. on Machine Learning (ICML)}, 2020.

\bibitem{Zhang2019Fixup}
H.~Zhang, Y.~N. Dauphin, and T.~Ma, ``Fixup initialization: Residual learning
  without normalization.'' in \emph{Int. Conf. on Learning Representations
  (ICLR)}, 2019.

\bibitem{he2016residual}
K.~He, X.~Zhang, S.~Ren, and J.~Sun, ``{Deep Residual Learning for Image
  Recognition},'' in \emph{IEEE Conf. on Computer Vision and Pattern
  Recognition (CVPR)}, Jun. 2016, pp. 770--778.

\end{thebibliography}

\end{document}